# No Keyword is an Island: In search of covert associations

Václav Cvrček[1], Masako Ueda Fidler[2]

> "The Democrats don't matter.
> The real opposition is the media.
> And the way to deal with them
> is to flood the zone with shit."
> Steve Bannon

**Abstract**: This paper describes how corpus-assisted discourse analysis based on keyword (KW) identification and interpretation can benefit from employing Market basket analysis (MBA) after KW extraction. MBA is a data mining technique used originally in marketing that can reveal consistent associations between items in a shopping cart, but also between keywords in a corpus of many texts. By identifying recurring associations between KWs we can compensate for the lack of wider context which is a major issue impeding the interpretation of isolated KWs (esp. when analyzing large data). To showcase the advantages of MBA in "re-contextualizing" keywords within the discourse, a pilot study on the topic of migration was conducted contrasting anti-system and center-right Czech internet media. was conducted. The results show that MBA is useful in identifying the dominant strategy of anti-system news portals: to weave in a confounding ideological undercurrent and connect the concept of migrants to a multitude of other topics (i.e., flooding the discourse).

# 1. Introduction

Many corpus-based discourse analyses start with a list of prominent units called keywords (KWs). However, as Scott aptly states, "KWs are just pointers" (Scott, 2010), devoid of precisely the contextual information which is crucial for determining the function or meaning of any linguistic unit (cf. "the meaning of a word is fully reflected by its contextual relations" Cruse, 1986: 16). After harvesting KWs, researchers must therefore find some way to explain the prominence-keyness of these words in their use in a corpus (e.g. Baker, 2006; Baker & McEnery, 2005). For this purpose several methods are normally used: e.g. close reading of concordances of KWs, collocation or cluster analyses of KWs, and/or examinations of links among KWs. These post-KW-extraction methods are valuable and informative, but not without issues, in disambiguating the pragmatic meaning and function of KWs.

Concordance analysis is highly interpretative and labour-intensive when dealing with large data, especially for frequently-occurring KWs. Examining collocations of KWs (or ngrams, clusters and other types of multi-word expressions) typically focuses on words in immediate or close proximity. Its results primarily pertain to the level of lexical (systemic) characteristics, rather than to the level of discourse. As topic-related words may not necessarily be in proximity, collocation analysis is

---

[1] Institute of the Czech National Corpus, Faculty of Arts, Charles University, vaclav.cvrcek@ff.cuni.cz
[2] Department of Slavic Studies, Brown University, masako_fidler@brown.edu

suboptimal for identifying associations between topics. KW links overcome this disadvantage by identifying KW co-occurrence within larger context windows (e.g.up to 15 words on either side). Despite their seeming similarity to MBA, however, KW links have no apparatus to evaluate their importance or strength; consequently, researchers cannot identify targets for closer analysis out of numerouos KW links

This paper proposes a new Market Basket Analysis (MBA)-based method to employ after KW extraction.MBA is a data mining technique which was originally used in marketing to identify which items customers are likely to buy together and use this to suggest their next purchases (e.g. "customers who bought X also bought Y"). This method, applied to texts (or, more precisely, the sets of KWs which characterize each text), sheds light on two important aspects of discourse, by which we we mean the overarching ideas that permeate a given set of texts; this definition is conceptually closest to that of Reisigl and Wodak (2016: 27). First, it produces objective evidence for associative networks of concepts that characterize the discourse of the whole corpus. Second, by revealing how concepts tend to regularly co-occur, MBA can be used to probe the implicit discourse agenda that may underlie the repetitive associative links (AL). The characteristics of a text can also be reflected in the patterns of co-occurring keywords . This co-occurrence is triggered by shared communicative or discursive functions. For example, Scott and Tribble (2006: 58) note that a "recipe for cake may well have several mentions of *egg*, *sugar*, *flour*, *cake*", and thus each would be identified as a keyword because of its high frequency. It is only through the co-occurrence of all those keywords in a text that researchers can conclude that the text is a recipe for a cake. The occurrence of only one would not lead to such a definitive conclusion. Thus it is the co-occurrence of keywords within a text that facilitates the identification of the text's aboutness and function. What is crucial here is that within the traditional keyword analysis these connections among KWs were almost always based on researchers' own judgment. Some connections among KWs, especially when their relations are unexpected or more subtle, may therefore escape notice.

The present study argues that MBA can enhance the quantitative analysis of texts where keyword analysis leaves off. We will demonstrate the merits of this method by contrasting data from the so-called "anti-system" online media class and data from the "mainstream" online media class in Czech. The results from MBA help reveal the mechanism of Bannon's reference to "flood[ing] the zone."

The following section describes the essential ingredients of MBA (section 2) as a method to enhance existing KW analysis. Section 3 presents the data used for the pilot study. Discussion of the results (section 4) is followed by conclusions (section 5).

# 2. KWA and MBA

Keyword analysis (KWA, Culpeper & Demmen, 2015) represents an established methodological framework, which is used for many purposes, from studies of literary texts (Culpeper, 2002; Scott & Tribble, 2006; Walker, 2010), to political speeches (Fidler & Cvrček, 2015, 2017), or in the analysis of public discourse on major social issues (Baker, 2005; Baker & McEnery, 2005; Tabbert, 2015, Fidler & Cvrček, 2018). This section discusses how to advance the study of KWs by a new strategy (MBA) that can be applied after KW extraction.

## 2.1 Preliminaries: KWA and frequency, dispersion, and context

This section describes the necessary steps that precede MBA. This discussion is essential because MBA as a post-KW-extraction method takes KWs as its input; the measures taken for their extraction can therefore significantly influence the results of MBA. There are three important pieces of information that have the potential to contribute to a comprehensive analysis of discourse based on KWs: frequency, dispersion, and context.

At the core of the KWA method is the identification of keywords (KWs), prominent units, in the target text or corpus against the backdrop of a reference corpus. Prototypical keyword analysis uses **frequency** as one of the key characteristics of KWs. In fact, the original operationalization of KWs is such that KWs are units with unexpectedly high frequency. Many analyses, moreover, do not address the internal structure of the corpus and treat all units included as a bag of words (see figure 1); the frequencies of words A to D below are all measured together regardless of the texts in which they occur. A rare exception to both of these methodological decisions is represented by Egbert & Biber (2019) with their proposal of dispersion keyness.

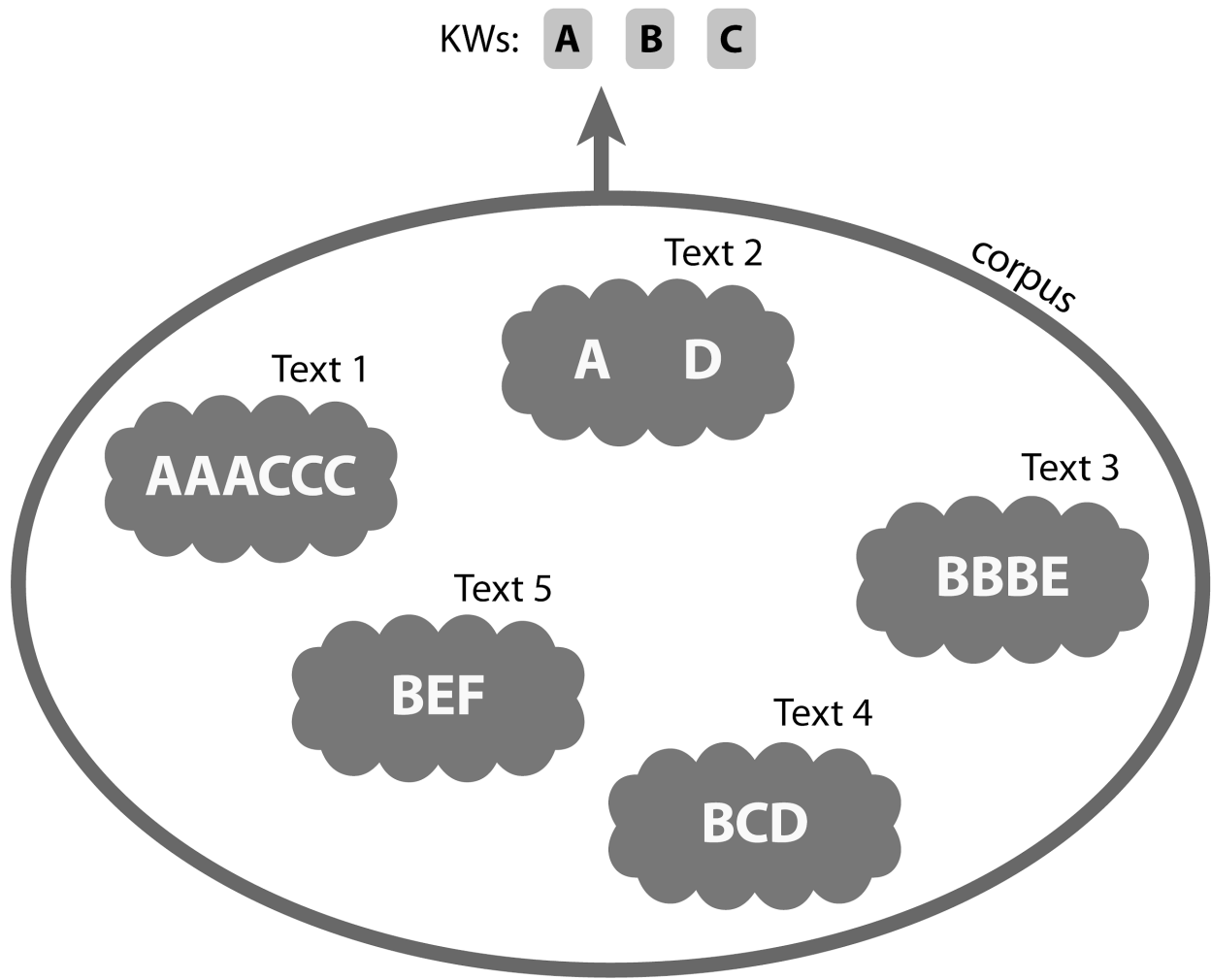

**Figure 1**. Schematic representation of KW extraction on a corpus consisting of multiple texts all treated together in a "bag-of-words" style.

The level of surprise can be evaluated by statistical significance tests, such as log-likelihood, chi2 or Fisher exact tests (cf. Bertels & Speelman, 2013), which compare the frequency of each word in the target text with the frequency of the same word in the reference corpus. Many KW lists are sorted according to p-value, assuming that it reflects keyness. However, it has been proposed several times (e.g. Hofland & Johansson 1982, Garielatos and Marchi 2012) that better evaluations can be achieved by inspecting effect size estimation, such as the Difference Index (DIN), a ratio (multiplied by 100) of the difference between relative frequencies (ipm) of an item in the target text (Ttxt) and the reference corpus (RefC) and the mean of those relative frequencies (Fidler & Cvrček, 2015):

$$DIN = 100 \times \frac{\text{RelFq}(Ttxt) - \text{RelFq}(RefC)}{\text{RelFq}(Ttxt) + \text{RelFq}(RefC)}$$

Values of DIN range from -100 to 100, where the extremes represent situations in which a word is missing either from the target text or the reference corpus, respectively. A DIN value of 0 indicates that word occurs in the target text and reference corpus with the same relative frequency.[3]

The fundamental role of frequency was recently challenged by Egbert & Biber (2019) and their proposal to use dispersion (or, more precisely, number of texts containing a word) instead. We acknowledge that one of the major disadvantages of the traditional "corpus frequency keyness" approach is that it treats the corpus as a single homogeneous unit (Egbert & Biber 2019: 78). Frequency, however, cannot be entirely disregarded in measuring keyness for two reasons. First, dispersion is a derivative of frequency; the calculation of all dispersion indices or adjusted frequencies known to authors (and listed by Gries 2008) presupposes token frequency. This is also true for the measure used by Egbert & Biber, which is described in its general form by Gries (2008: 407) as *range* (i.e., the number of corpus parts containing a word at least *x* times, i.e. having the frequency ≥ *x*). Second, the use of frequency to capture prominence is not only justified, but is the only viable option under certain research scenarios. For example, use of frequency is indeed justified when a target corpus consists of a single homogeneous unit (text). Frequency must be considered when the target corpus cannot be split into meaningful parts – an operation needed to calculate dispersion based on segments of text/corpus (e.g. the analysis of one newspaper article which is too small to be cut into segments). The latter situation applies to the data used in this paper.

In a research scenario where we examine a corpus of predominantly short texts (typically, articles in a newspaper within a specific time span) we would also like to have control over the KW **dispersion**. A typical KW extraction, as shown in figure 1, involves harvesting a "bag of words". However, the prominence of a word per se does not necessarily mean that it is widespread in the texts within the corpus. One way to integrate dispersion is therefore to analyze a list of *key keywords* (Scott & Tribble, 2006, p. 77) by identifying KWs that occur in multiple texts of the target corpus. The other way is to alter the raw (or token) frequency of a word in the corpus with the number of texts in which it occurs (Egbert & Biber, 2019). Both of these methods help to correct the drawbacks of the "bag of words" approach.

The third ingredient is more complicated than the preceding two: accessing the **context** in which a KW occurs. A word (regardless of its prominence) which is "flying in a vacuum" is of limited use to linguists, especially in discourse analysis (cf. "collocations create connotations" Stubbs, 2005: 14). KW extraction is therefore normally followed by additional methods: e.g. close reading of concordance lines (Egbert & Baker, 2019: 56), collocation analysis of KWs (e.g. Gabrielatos and

---

[3] The idea of effect size to weigh prominence is by no means a recent discovery. It goes back to Hofland and Johansson's Difference Coefficient (Hofland & Johansson 1982). For discussion on DIN and other formulae to assess effect size, see Fidler and Cvrček 2015.

Baker 2008), key multi-word expressions, ngrams or clusters (Fischer-Starcke, 2009; Mahlberg, 2007; Partington & Morley, 2004), or examining links between KWs (Scott & Tribble, 2006: 73) within a larger span of words (e.g. circa 15 words). These existing methods are valuable tools to study the function of a KW within a local context.

KWs, however, may co-occur across a much longer distance, e.g. across paragraph boundaries within the same text. For instance, the KWs *ship* and *summit* may not necessarily occur in close proximity. This type of co-occurrence informs us of the conceptual network (or a narrative) *of the entire text*. If this co-occurrence pattern is found repeatedly in multiple texts, we may consider it a signal of consistent conceptual association that is part of an overarching narrative of the entire corpus. This effect can be viewed as a discourse correlate of collocation: collocation reveals that the meaning of the recurring word *illegal* in the vicinity of *immigrant* becomes part of the image of the latter (Stubbs 1996: 197); in a similar way, if KWs *EU, countries, vassal, obedient* recur together in **multiple** texts, they become part of a narrative of the entire corpus (e.g. "countries as obedient servants of the EU"), i.e. part of the ideological underpinning typical of a certain corpus.

The need to objectively account for the contexts in which KWs occur and their relations to one another brings us to Market Basket Analysis (MBA), which is the main focus of this paper. MBA, with its algorithmic and reproducible method of identifying repeating patterns of KW co-occurrence in texts of the corpus, can extend the current quantitative corpus-driven methods to facilitate qualitative interpretation.[4]

## 2.2 Market basket analysis: how it works for text analysis

Market-basket analysis (MBA) is a data mining technique to identify frequent patterns in a data set (Han et al., 2011: 244; Miner, 2012: 917): e.g. it identifies items that are frequently bought together in one transaction. Recurrent purchases of various items within one "shopping cart" reflect the customer's shopping behaviour; such items are considered as "being associated," forming an associative link (AL). MBA is capable of identifying recurring associative relations by sifting through large amounts of transactions. MBA thus captures some important aspects of who the customers are or points to their motivations and intentions. Similarly, in discourse analysis, MBA reveals tendencies of KW(s) to co-occur with others; in other words, MBA indicates how certain concepts gravitate together and create connections (see figure 2).

---

[4] Cf. "While the earliest stages of a corpus analysis tend to be quantitative, relying on techniques like keywords and collocates in order to give the research a focus, as a research project progresses, the analysis gradually becomes more qualitative and context-led, relying less on computer software." (Baker & McEnery, 2015: 2)

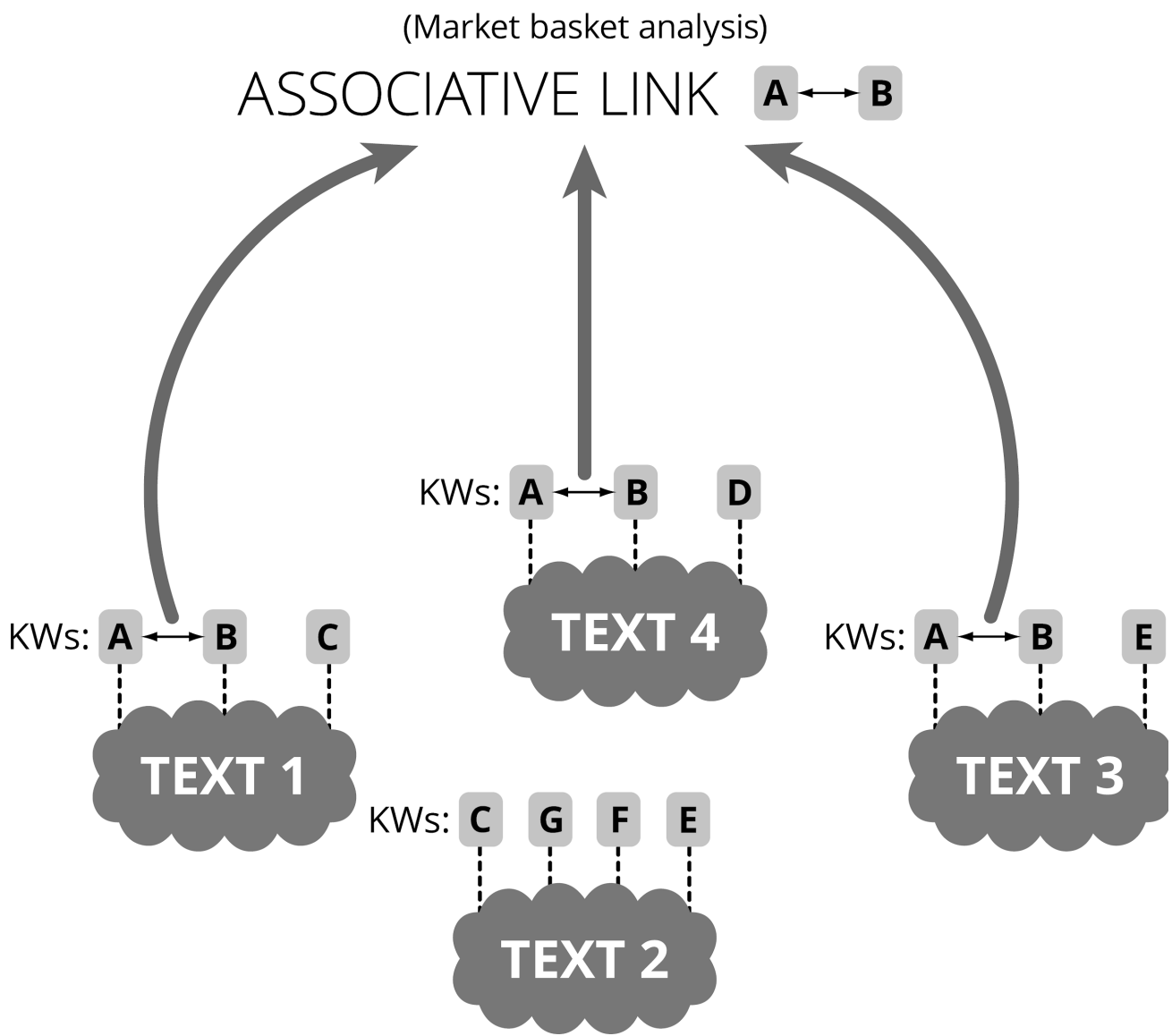

**Figure 2.** Schematic representation of KW extraction from individual text and establishing associative links with MBA

As mentioned above, MBA's inputs are lists of KWs identified in *texts* (a separate list for each text). The algorithm then evaluates each possible combination of KWs (pairs, triplets… up to a size specified by the user), yielding link candidates. As the list of these link candidates can grow enormously large, uninteresting ones were filtered out by three variables: support, confidence and lift. These variables help to select out noteworthy combinations of KWs: those which are attested in a certain percentage of texts; those whose constituents' co-occurrence is above a certain threshold; and those above a certain amount of associative strength. Support and confidence are used as thresholds below which any combination is discarded. Lift is used for sorting the resulting list according to the associative strength. A brief mechanism of these three measures is described below.

Suppose we have *N* number of texts in a corpus and for each one of them we have identified a set of KWs. One of these KWs is *A*. *Support* for each KW is defined as a proportion (or probability) of texts containing *A* as a KW.

$$\text{support}(A) = \text{texts}(A)/N = p(A)$$

Now, let us consider the association between two KWs: *A* and *B*. The associative link *A* ➙ *B* consists of two KWs in this case: the left-hand side (or lhs) of the link (*A*) is called the antecedent and the right-hand side (rhs) of the link (*B*) is the consequent . The support of the associative link *A* ➙ *B* is the proportion of texts which contain both *A* and *B* as KWs (or the spread of the link *A* ➙ *B* within the corpus):

$$\text{support}(A \to B) = \frac{\text{texts}(A, B)}{N} = p(A, B)$$

For example, support for the associative link (*migrant* ➙ *EU*) within the center-right web portals is calculated as the number of texts containing both words as KWs (which is 110 in our data, see section 3) divided by the total number of texts analyzed in this segment (12,110), i.e. 0.0091.

*Confidence*, the second measure, is a conditional probability which represents how often *B* appears in texts that contain also *A*.

$$\text{confidence}(A \to B) = \frac{\text{texts}(A, B)}{\text{texts}(A)} = \frac{p(A, B)}{p(A)}$$

The confidence for the same AL (*migrant* ➙ *EU*) in center-right media equals 110 (i.e. number of texts containing both *migrant* and *EU* as KWs) divided by 259, which is the total number of texts containing *migrant* as KW: 110/259 = 0.425.

Support and confidence measure the reliability of the associative link. When counting the number of texts containing a given KW, support reflects the range of the link (in the sense used by Gries, mentioned above) with respect to the size of the corpus. Confidence, on the other hand, reflects the range of the link with respect to the frequency of the antecedent. Support of 10 % means that the association under examination (*A* ➙ *B*) occurs in 10 % of texts in the *entire (sub)corpus*. Confidence of 50 % means that *B* is obligatory in half of the *texts containing A* – in other words, 50 % of texts containing *A* are instances of the associative link *A* ➙ *B* (cf. Han et al., 2011: 245).

Given the input variables of support and confidence (the total number of texts *N*, the number of texts containing *A*, the number of texts containing both *A* and *B*), both measures clearly reflect the relative dispersion of KWs in the corpus (in a similar way as is calculated by Egbert & Biber 2019).

*Lift*, the remaining variable, represents how much our confidence has increased that *B* will be present in a text given that *A* is already in the text.

$$\text{lift}(A \to B) = \frac{p(A, B)}{p(A) \times p(B)} = \frac{\text{texts}(A, B) \times N}{\text{texts}(A) \times \text{texts}(B)}$$

Lift for the associative link (*migrant* ➙ *EU*) uses the above mentioned figures – the number of texts containing both words as KWs (110), the number of texts containing *migrant* as KW (259), and the total number of texts in the center-right segment (12,110) – and the number of texts containing *EU* as KW (which is 568). The lift value is thus (110 × 12110) / (259 × 568) = 9.055.

Thus, lift can be conceptualized as the number of texts in which KWs *A* and *B* co-occur divided by the product of the number of texts containing either word. When transformed to probabilities of occurrence, the formula for lift calculation can be proved to be conceptually similar to the mutual information *I*, which is at the core of the association measure MI-score (Church & Hanks, 1990) used for collocation extraction. Thus the associative links are justified as a higher-level correlate of collocation (2.1).

Among the three measures, lift is the indicator for the strength of the association. While support and confidence are used as thresholds to filter out associations that do not affect the decisive proportion of texts (cf. Information Resources Management Association, 2014: 1090), lift is used to rank the associations between KWs. The mathematical principle behind lift parallels MI-score in collocation lists and keyness measures in KW lists, both of which are trustworthy methods of ranking various types of strength.

As there is no previous experience with these measures, the thresholds had to be established empirically by primarily taking into account the necessity of keeping the total number of MBA-generated associative links to a manageable quantity. Future research should therefore seek to investigate the pitfalls and limitations of particular MBA settings.

MBA as a post-KW-extraction procedure has the potential to solve the issues inherent in the approaches mentioned in the introduction[5], as it is built on the information about repeated co-occurrence of KWs in multiple texts, in addition to the usual concept of token-frequency based KWs. The relative proportion of texts containing both KWs in a link – as reflected by support – guarantees a certain level of dispersion across the target corpus. By identifying recurrent links among KWs, MBA fulfills the requirements of repeated occurrence of a set of KWs in multiple texts and provides the broader (discourse-level) context (associations to other KWs). Moreover, it has a built-in and well-established means to evaluate associative strength (lift), thereby allowing researchers to assess what types of narrative are predominant within the entire corpus even when it contains large amounts of data. The following sections will show how MBA might be applied to the specific WebMedia corpus.

# 3. Methodology and data

This section presents a pilot study to show how MBA can contribute to discourse analysis after KW extraction. The exploratory study was conducted on theWebMedia corpus (80 million words and 94 million tokens, incl. punctuation), which consists of Czech online web media portals from October 2017 to October 2018.

WebMedia contains 211,000 media articles in total. The data, obtained from targeted web-crawling, reflects a large spectrum of Czech web portals ranging from mainstream (center-right, center-left) to anti-system or alternative media. The corpus is a byproduct of a larger ongoing project to develop an infrastructure for a large web-crawled corpus (called ONLINE) which is updated on a daily basis and covers not only a wide range of web media but also social networks, discussions, and web forums in Czech.

The WebMedia corpus is divided into 6 major media classes: alternative, anti-system, tabloids, center-left mainstream, political tabloid, center-right mainstream (Table 1):

| Media class | Tokens | Articles |
|---|---:|---:|
| Alternative (*alternativní*) | 3,761,057 | 4,556 |

---

[5] It is not our aim, however, to argue that MBA replaces these methods. Collocation analysis, KW analysis and other methods mentioned here could be used *after* MBA when it is necessary to focus on word usage on a more local level of discourse.

| | | |
|---|---:|---:|
| Anti-system (*antisystémové*) | 14,666,404 | 19,031 |
| Tabloids (*bulvár*) | 13,435,010 | 44,059 |
| Center-left mainstream (*levý střed*) | 14,511,164 | 36,900 |
| Political tabloid (*politický bulvár*) | 29,617,511 | 69,122 |
| Center-right mainstream (*pravý střed*) | 18,121,533 | 37,769 |

**Table 1.** Media classes in the WebMedia corpus.

The media classification used above is drawn from an audience-based typology of Czech online news portals devised by Josef Šlerka (*Mapa médií* 'Media map', see http://www.mapamedii.cz/mapa/typologie/index.php). *Mapa médií* groups the media types according to the shared preferences of their readers.[6] It is essentially informed by reader behaviour from two services: *Facebook* (activities such as likes and shares of papers/portals) and *Alexa Rank* (https://www.alexa.com/, a service providing information about people's behaviour on the web for the purpose of market research).

The media are thus not classified by their linguistic characteristics, topic preferences or political stance; portals clustered in one category have a significant overlap in audience. The only part of the classification that is subject to researcher's interpretation is the labeling of clusters (the media class labels in Table 1), which is derived from the central items of each category (prototypical portals in each type). *Mapa medií* rules out subjective judgment on the classification of media which is important, especially in the Czech media landscape where the media cannot be easily classified along established categories as e.g. tabloids vs. broadsheets in the UK.

The WebMedia corpus is lemmatised and morphologically tagged with up-to-date tools (Jelínek 2008, Petkevič 2014), which are also developed and used for corpora of written Czech within the Czech National Corpus (CNC) project. These tools allow KW extraction with SYN2015 (Křen et al. 2015) as a reference corpus, without the risk of introducing interferences caused by incompatible annotations (namely alternative approaches to lemmatization). SYN2015 is a 100m representative corpus of contemporary written Czech compiled within the CNC project; for detail about its composition see Křen et al. 2016.

We analysed the data of anti-system (ANTS) and center-right mainstream (CR) portals in two steps (a schematic depiction of the KWA–MBA interface is in Figure 2).

1. A set of KWs for each article within ANTS and CR portals was obtained (with SYN2015 as a reference corpus); this step yielded more than 19,000 KW lists for ANTS and almost 38,000 lists for CR (the number of lists was reduced eventually, see below).
2. MBA was applied to the results from step 1.

Below we present the details of the two steps. KWA was carried out with the following settings.

We used a log-likelihood test (Dunning 1993) for comparing the relative frequencies in the target and reference corpus and for filtering out statistically insignificant candidates for KWs, but we also paid attention to the effect size which we measured with DIN (Fidler & Cvrček 2015). The analysis was carried out on lemmas for two reasons: first, we wanted to focus on concepts rather than morphological properties and, second, we wanted to avoid low frequency items by consolidating numerous word-forms, e.g. *žena* 'woman (Nom. sg.)', *ženě* 'to/about a woman (Dat. or Loc. sg.)', *ženama* 'with women (Inst. pl.)' under one lemma *žena* 'woman'.

---

[6] The classification adopted in the WebMedia corpus here is based on the original version of the media typology available at the time of WebMedia corpus compilation. The newer version available at the *Mapa medií* web site as of March 2021 (more fine-grained, with slightly different labels) is used for the publicly available ONLINE corpus (see https://wiki.korpus.cz/doku.php/en:cnk:online).

- The statistical significance was measured by the p-value of the log-likelihood test; the threshold we applied was set on a confidence level α = 0.001 (which corresponds to the log-likelihood test statistic value of 10.83).
- The effect size was measured by DIN > 70 (i.e. only words which are overrepresented in the target texts, see 2.1).
- The minimum frequency of a word in the target text was set to 3 and only words written with alphabetic symbols were considered.

These thresholds represent a conservative setting of KWA in Czech, which slightly prioritizes KWs recall over precision. In this setup we utilized our experiences with previous analyses focusing on prominence on various linguistic levels (Fidler & Cvrček, 2018, Cvrček & Fidler 2019).

The output of keyword extraction (list of KWs for each article) was then processed by MBA. However, not all texts were analyzed by MBA. Very short texts with only a few dozen words did not provide a reasonable surface area to study associations between concepts. Thus, a decision was made to apply MBA only to those KW lists containing 15 or more KWs (the value was set up empirically) as the main goal is to analyze co-occurrence of KWs. Having applied this criterion, the input of MBA consisted of 7,401 KW lists in ANTS (39 % of the total number of texts in this segment) and 12,110 lists in CR (32 % of the total number of texts in this media class). As a consequence of this reduction the average length of texts in the material has doubled (in ANTS it has risen from 770.7 tokens in the whole corpus to 1375 in the reduced data, and in CR it has risen from 479.8 in the whole WebMedia corpus to 862 tokens in the data analyzed through MBA).

Mining of associative links (ALs) by MBA may be very time-consuming because of the large number of texts under consideration (and corresponding number of KWs in them) and the need to inspect each combination of KWs (for description of the procedure see 2.2). Therefore a powerful algorithm with sophisticated pruning of the decision tree must be used. In our study we used the *Apriori* algorithm (Agrawal & Srikant, 1994), which was implemented in package *arules* (Hahsler et al., 2011, 2019) in R (R Core Team, 2018).

MBA was carried out with the following specifications:

- the minimum value of support was set at 0.003;
- the minimum value of confidence was set at 0.4.

In other words, only those combinations of antecedent–consequent occurring at least in 0.3% of texts in each corpus were considered, and the minimum relative co-occurrence of antecedent and consequent in these texts was set to 40 %. The values of both thresholds were assigned empirically, resulting in a reasonable amount of ALs in both subcorpora.

The last parameter of MBA, length, is of crucial importance because it states how many KWs can be involved in the process of establishing an AL. We set this parameter to 4, a setting that allows three types of links (with respect to the size of antecedent):

A ➙ X

A, B ➙ X

A, B, C ➙ X

In other words, length controls the KW combinations in the antecedent and so the size of the link. An example of the last type of link is *american*, *north*, *Trump* ➜ *Korea*, which can be interpreted as showing a tendency to include the KW *Korea* in the same text where *american*, *north* and *Trump* are already present as KWs. After trying out several other values of length, 4 was deemed optimal with respect to the trade-off between the number of ALs identified and the number of overlaps between them (for each link of the size *n* there are up to *n* links of the size *n-1* etc.).

With the settings described above, we obtained 284,767 ALs for anti-system (out of 7,401 analyzed texts) and 60,359 ALs for center-right mainstream (out of 12,110 analyzed texts).

In the setup for MBA, attempts were made to keep the threshold values as permissive as possible (to maximize the recall of associative links), and at the same time to keep the total number of extracted links manageable for further interpretation. The potential caveat is that the settings for MBA was empirically rather than linguistically motivated. Here, it is necessary to point out that this study is the first attempt to apply MBA to text analysis. Its settings are therefore unlike the threshold settings for KW extraction for which the values have been tested and gradually conventionalized in a number of previous studies. The main focus of this paper is to propose a new potential methodological framework for extracting associations from KW lists; discussion on the effect of cutoff points is a topic for further research.

# 4. From associations to overarching narratives of *migrant* in Czech online media

This section examines ALs from two types of web media classes: center-right (CR) and anti-system (ANTS) media classes (Table 1). The data in the corpus were drawn from the time when refugees were arriving in Europe (especially from the Mediterranean through Libya) in large numbers (from October 2017 to October 2018). Czechs intensely followed these events in their media. This was also the time of heated discussions on revisions to refugee quota regulation (Dublin IV) and the Marrakesh declaration, which defends the rights of religious minorities in Muslim majority countries.

CR represents the most dominant Czech mainstream media class and ANTS a media class at the margins of the Czech web portals. The latter include e.g. *Sputnik Czech Republic* (https://cz.sputniknews.com), which is sponsored by the Kremlin and is known to disseminate pro-Russian views (Fidler and Cvrček, 2018, Cvrček and Fidler, 2019); some others (e.g. Aeronet or AENews, https://aeronet.cz/news/) are suspected of receiving funding from Russia.[7] ANTS are generally thought to show nationalist (anti-EU), pro-Russian, anti-semitic, and racist tendencies; apocalyptic views of the world are also typical of these portals (Fidler and Cvrček 2020).

The current analysis aims to demonstrate some marked differences between CR and ANTS with respect to ALs that include the KW *migrant*[8] in CR and ANTS. Although not as a comprehensive treatment of a similar topic as Baker et al. 2008 and Gabrielatos and Baker (2008), the present study shows how MBA helps to highlight differences in the overarching narratives (a more detailed study is in preparation by the authors). Links obtained by MBA were studied quantitatively (in 4.1) as well as qualitatively (by interpreting ALs and inspecting texts containing elements of the link in 4.2).

---

[7] https://www.irozhlas.cz/zpravy-domov/frantisek-vrabel-dezinformace-fake-news-rusko-facebook_1903130001_ogo

[8] *Migrant* in Czech should not be conflated with the English equivalent of "migrant," although there is some semantic overlap. The word refers to a migrating individual or a refugee (cf. Petráčková et al. 1998 or https://prirucka.ujc.cas.cz/?slovo=migrant); however, we find little evidence of *migrant* used in the sense of seasonal workers, which is found in the English definition of 'migrant' in e.g. Oxford English Dictionary.

## 4.1 Anti-system vs. Center-Right Media Classes: An overview

Table 2 and Figure 3 offer a glimpse into the possible overall features characteristic of each media class. The lemma *migrant* occurs 7196 times in the ANTS texts and 4211 times in the CR part of the corpus. However, as we decided to restrict the corpus to longer texts for the purpose of studying ALs, in the following we focused on 3450 occurrences in ANTS and 1671 occurrences in CR.

| Media class | number of texts with *migrant* as a KW | number of ALs | median lift | median support |
|---|---|---|---|---|
| CR | 256 (2.1 % of all CR texts) | 235 | 14.5 | 0.0038 |
| ANTS | 472 (6.4 % of all ANTS texts) | 1448 | 7.01 | 0.0039 |

**Table 2**. Overall numbers characterizing the ALs containing lemma *migrant* in ANTS and CR.

Table 2 shows the number of texts where *migrant* occurs as a KW and the number of associative links with this KW found in these texts. The lift values and the numbers of ALs point to distinct discourse properties of ANTS and CR. ANTS yields almost twice the number of texts containing *migrant* as a KW (472 vs. 256) and six times more ALs (1448) than CR (235) containing *migrant*. Note also the overall strong associative links with *migrant* in CR in contrast to ANTS (the difference in median lift values is statistically significant, $p < 0.0001$; Wilcoxon rank sum test); the interquartile range of lift in CR is wider than in ANTS (see Figure 3).

These results suggest much stronger preoccupation with *migrant* in ANTS by indicating more texts including the KW than in CR (in absolute as well as in relative numbers). The results also suggest that ANTS creates a wider and diffuse network of associations with *migrant* than CR. This pattern may reflect the strategy of "flooding" (as used by Bannon) with a large repertoire of loosely connected associations. ANTS, then, might "piggyback" on the topic of migration to expound on other topics; e.g. Islamization of Europe, racism, conspiracies about NGOs, the EU in disarray and its damaging influence on Czechia–staple themes that appear in ANTS in general. As for CR, despite the overall smaller number of ALs compared to ANTS, many of the ALs are strongly linked to *migrant*. It is possible that *migrant* is discussed in conjunction with a relatively small set of topics such as the EU parliament and its debates. This is explored in more detail in 4.2.

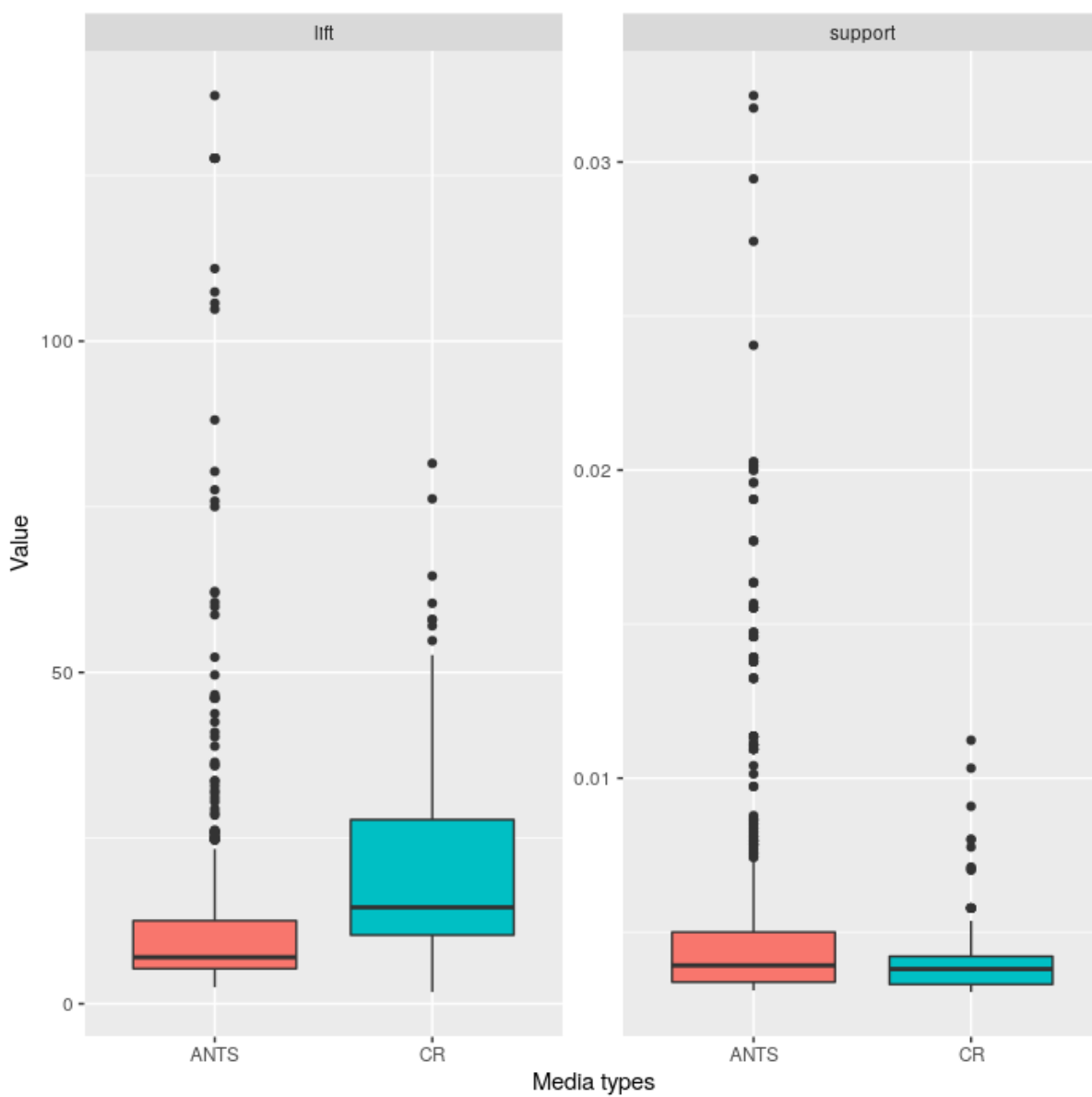

**Figure 3.** Values of lift and support (N.B. different y-axis scales) for ALs containing the lemma *migrant* in CR and ANTS.

Before turning to a more detailed examination of ALs, additional comments on support values must be made as they also add to the "flooded" nature of ANTS texts.

Indeed, ANTS and CR differ significantly in the support variable (p = 4.535e-05; Wilcoxon rank sum test) despite having almost identical medians (the extreme outliers should not affect the result, as the test is rank-based). The overall higher ANTS support values could be explained either (i) by the media's persistent preoccupation with the topic of migration (i.e. the media class is full of texts containing *migrant*) or (ii) by its intentional efforts to make an anti-migrant argument (that would eventually lead to anti-EU argument, see below) wherever possible (even in texts that are not directly related to migration). The former hypothesis can be cross-checked by measuring the overall thematic concentration (TC), a widely-used variable in quantitative linguistics that shows how diverse or compact (thematically) a text is.

Thematic concentration (Popescu et al. 2007) is derived from the "h-point" in the frequency list of a text or corpus. The h-point corresponds to the word in this list which frequency is equal to its rank (e.g., the 57th word in the frequency list of types in a text with a raw frequency of 57 occurrences). This h-point splits the distribution into two different parts: words with a frequency higher than the h-point (usually grammatical words) and all other words (usually lexical/content words). Words contributing to thematic concentration are those lexical- or content words which can be found above the h-point, which is populated by grammatical (functional) words.

We have calculated TC for both the ANTS and CR subcorpora. The results show that TC for CR (0.0052) slightly surpasses that for ANTS (0.0044). The higher value for CR suggests a higher concentration of texts on fewer topics and thus a smaller width of the topics covered by CR media while lower value of TC for ANTS confirms the suspicion (ii) that ANTS' strategy is partly flooding the discourse with diverse topics to which migrants are loosely linked.

MBA clearly shows some major differences between ANTS and CR. The following section takes a closer look at the ALs in CR and ANTS.

## 4.2 Salient features of ALs in CR and ANTS

This section discusses ALs in more detail to illustrate the differences between CR and ANTS, especially with repsect to the latter. To distinguish the two media classes, we looked at the ALs that are outliers[9] in terms of the highest lift values, which are expected to show the characteristic features of each media class. Since the number of outlier ALs is high, a special attention was paid to the ALs containing those KWs that are unique to the media class. When a unique KW is included in multiple ALs, the AL with the highest lift value was chosen for closer examination. For example, the KW *Italy* appears in more than one AL (Table 3): **Italian**, *migrant* → *Italy* and *ship, migrant* → *Italy*; the former AL with the higher lift value was chosen for closer examination.

Tables 3 and 4 show the outlier ALs (the English gloss) in each class, cf. Figure 3: ALs above 52.6 for CR and ALs above 23.4. The KWs in boldface are unique to the media class. The tables with the original Czech KWs are in Appendix 1.

### 4.2.1 Outliers in CR

The unique KWs in ALs in Table 3 mark CR's interest in negotiations within the EU summit to solve the migration crisis (*summit*), migrants arriving to Italy by sea, and Italian reaction to migrants (*Italy, Italian* → *ship*).

| ranking | lhs | rhs | lift | count |
|---|---|---|---|---|
| 1 | European (adj), migrant, **summit** | migration (adj) | 81.5 | 38 |
| 2 | migrant, **summit** | migration (adj) | 76.2 | 38 |
| 3 | **Italian** (adj), migrant | **ship** | 64.6 | 39 |
| 4 | EU, European, migrant | migration (adj) | 60.4 | 42 |
| 5 | European, migrant, country | migration (adj) | 58.0 | 46 |
| 6 | **Italian** (adj), migrant | **Italy** | 58.0 | 54 |
| 7 | **ship**, migrant | **Italy** | 57.0 | 42 |
| 8 | European (adj), migrant | migration (adj) | 54.8 | 56 |

**Table 3**. Outlier ALs that include *migrant* in center-right web media, ranked by lift. The "count" column shows the number of texts in which the AL is attested.

Each AL can be found in a number of texts (see the last column of Table 3). The interpretation of the link provided below was drawn from close reading of the texts in which it appears. However, due to space limitations we were able to include only the title of one article or a short excerpt of a text to exemplify the narrative that the AL represents (see Appendix 1 for the Czech originals as well as their URLs).

---
[9] The list of all KWs and all ALs identified in the data is available at https://osf.io/3xjp9/.

AL *European, migrant, summit → migration* can be exemplified by an article "No one wants the ship with migrants, Merkel did not appease the rebels with the summit" (Appendix 2-1), which represents Angela Merkel as a leading figure in the EU and the opposing parties as "rebels". The article reports on the EU debates that did not bring conclusive results. Texts with ALs including *Italy* present Italy's reaction to the crisis and a growing support for populists in Italian elections; for example, AL *Italian, migrant → ship* is in an article "Another ship with migrants set out [to the sea]. Italy rejected it, it will finally anchor in Spain" (Appendix 2-2); or AL *Italian, migrant → Italy* is found in "Italian elections at the signal of migration. Populists are leading in the polls, promising to stop it, but they do not challenge the EU or the eurozone" (Appendix 2-3).

Examination of outlier ALs with unique KWs, followed by inspection of texts that contain these links, helps us understand the overarching narrative on migrants and migration based on one KW *migrant* and its strongest associated KWs: the migration crisis is presented as a challenge to the EU; while there are dissenting voices (the "rebels"), exit from the EU is not proposed as a solution. Outlier ALs do not include KWs related to the details of rescue efforts or the support and integration of migrants; this suggests that CR maintains a detached picture of the migration crisis as a problem to be resolved among the EU member states.

### 4.2.2 Outliers of ANTS

The first impression of the outliers in ANTS points to different stories. Table 4 clearly shows not only the larger number of ALs and a wider range of lift values as outliers, but also a larger variety of KWs in ANTS than in CR. These results are consistent with the suggestion in 4.1 that ANTS might flood their texts with various loosely connected concepts in association with the KW *migrant* (original Czech KWs are in Appendix 1).

| ranking | lhs | rhs | lift | count |
|---|---|---|---|---|
| 1 | **Dublin**, **migration**, migrant | IV | 137.1 | 28 |
| 2 | **IV**, migrant | Dublin | 127.6 | 34 |
| 3 | EU, **IV**, migrant | Dublin | 127.6 | 29 |
| 4 | **IV**, **migration**, migrant | Dublin | 127.6 | 28 |
| 5 | European, IV, migrant | Dublin | 127.6 | 27 |
| 6 | IV, migrant, country | Dublin | 127.6 | 26 |
| 7 | **Dublin**, migrant | IV | 111.0 | 34 |
| 8 | **Dublin**, EU, migrant | IV | 107.4 | 29 |
| 9 | **Dublin**, European (adj), migrant | IV | 105.7 | 27 |
| 10 | **Dublin**, migrant, country | IV | 104.8 | 26 |
| 11 | migrant, **applicant**, country | asylum | 88.1 | 23 |
| 12 | migrant, **applicant** | asylum | 80.3 | 31 |
| 13 | **asylum**, migrant | applicant | 77.6 | 31 |
| 14 | **asylum**, migrant, country | applicant | 75.9 | 23 |

| 15 | **Africa**, migrant, country | **African (adj)** | 75.0 | 26 |
| 16 | **Africa**, EU, migrant | **African (adj)** | 62.2 | 23 |
| 17 | **Hungary**, migrant | **Hungarian (adj)** | 62.0 | 23 |
| 18 | **Africa**, **Europe**, migrant | **African (adj)** | 60.6 | 28 |
| 19 | **Africa**, **migration**, migrant | **African (adj)** | 59.9 | 24 |
| 20 | **Hungarian (adj),** migrant | Hungary | 58.7 | 23 |
| 21 | **Africa**, migrant | **African (adj)** | 52.3 | 29 |
| 22 | **African (adj)**, **Europe**, migrant | **Africa** | 49.6 | 28 |
| 23 | **document**, **migration**, migrant | **legal** | 46.6 | 24 |
| 24 | **declaration**, migrant | **legal** | 46.1 | 23 |
| 25 | **declaration**, **migration**, migrant | **legal** | 46.1 | 23 |
| 26 | **African (adj)**, migrant, country | **Africa** | 46.1 | 26 |
| 27 | **African (adj)**, EU, migrant | **Africa** | 43.8 | 23 |
| 28 | **African (adj)**, **migration**, migrant | **Africa** | 42.5 | 24 |
| 29 | **migration**, migrant, UN | **legal** | 41.0 | 29 |
| 30 | **African (adj),** migrant | **Africa** | 40.3 | 29 |
| 31 | **document**, migrant | **legal** | 38.9 | 25 |
| 32 | migrant, **refugee**, country | **asylum** | 36.5 | 24 |
| 33 | migrant, **UN** | **legal** | 36.1 | 29 |
| 34 | **migration**, migrant, **politics** | migration (adj) | 35.9 | 24 |
| 35 | **migration**, migrant, **refugee** | **legal** | 33.7 | 26 |
| 36 | **CzR**, migrant, country | **member (adj)** | 33.6 | 28 |
| 37 | **legal**, **migration,** migrant | migration (adj) | 32.9 | 25 |
| 38 | **commission**, migrant, country | **member (adj)** | 32.2 | 23 |
| 39 | **Africa**, migrant, country | **legal** | 31.8 | 23 |
| 40 | **Afrika**, **migration,** migrant | **legal** | 31.1 | 26 |
| 41 | **migration,** migration (adj), migrant | **legal** | 30.5 | 25 |
| 42 | migrant, **Germany**, **German (adj)** | **Merkel** | 29.4 | 30 |
| 43 | EU, **commission**, migrant | **member (adj)** | 28.8 | 24 |
| 44 | migrant, **our**, country | **member (adj)** | 28.5 | 26 |
| 45 | migrant, **minister** | **Interior (noun)** | 26.2 | 25 |
| 46 | **CzR**, EU, migrant | **member (adj)** | 26.1 | 28 |
| 47 | **Africa**, **Europe**, migrant | **legal** | 25.9 | 25 |
| 48 | **declaration**, migrant | **migration** | 25.9 | 31 |
| 49 | European, migrant, **UN** | **migration** | 25.9 | 31 |
| 50 | **legal**, migrant, **UN** | **migration** | 25.9 | 29 |

| 51 | **declaration**, **legal**, migrant | migration | 25.9 | 23 |
| --- | --- | --- | --- | --- |
| 52 | **global**, migrant, **OSN** | migration | 25.9 | 23 |
| 53 | migrant, **German (adj)** | Merkel | 25.2 | 33 |
| 54 | **Evrope**, **legal**, migrant | Africa | 25.2 | 25 |
| 55 | European, **commission**, migrant | member (adj) | 25.2 | 24 |
| 56 | **Africa, legal** migrant | migration | 25.0 | 26 |
| 57 | **document, legal**, migrant | migration | 24.8 | 24 |
| 58 | **Africa**, migrant, **union** | migration | 24.8 | 24 |
| 59 | EU, migrant, **NGO** | migration | 24.8 | 24 |
| 60 | migrant, **NGO**, country | migration | 24.8 | 23 |
| 61 | migrant, **UN, refugee** | migration | 24.8 | 23 |
| 62 | **migration**, migrant, **union** | Africa | 24.7 | 24 |

**Table 4.** Outliers in anti-system web media that include *migrant*, ranked by lift. in the anti-system web media. The "count" column shows the number of *texts* in which the AL is present.

The ALs with very high lift value (above 100) indicate references to the Dublin IV discussion on the redistribution of migrants. The ALs with lift above 58 include KWs related to Hungary, a country that has opposed the redistribution of migrants. KWs referring to Africa and legality [of migrants] appear in many ALs; in fact, the negated form *nelegální 'illegal'* of the lemma *legální* occurs much more frequently (1345) than the positive counterpart (642). The KW *migration*, a nominalization of an action, occurs in different constellations. Outlier ALs towards the bottom of the table contain a large variety of KWs: those related to government officials and politics (*minister, interior, politics*), those related to agreements and mechanisms of accepting migrants (*refugee, document, declaration*), those that hint at domestic affairs and national interests (*Czech Republic*, the possessive pronoun *our*), those that refer to the key EU players and the neighbor of the Czech Republic (*Germany, Merkel*), and those indicating transnationality (*UN, NGO, global*).

By examining the texts including these links, several "narrative lines" were identified, including negative images of migrants, the Czech government's collusion with the EU, the EU as an authoritarian system and a need for Czech EU-exit, the mainstream media hiding the truth, and global conspiracies by transnational organizations. Below are stories that can be extracted from ANTS texts. Each narrative line is discussed separately below.

**Negative images of migrants**
One negative image of migrants occurs with the KWs *Africa* and *African*, which are often used as meronyms for migrants, suggesting a sweeping stereotype for all migrants. In fact, the overgeneralization of migrants' identities as Africans and Muslims (yet another aspect of stereotyping) and as "violent" and "parasitic" can be observed in texts with AL *Africa, migrant, country → African*: "Afroislamic violent migrants and economic parasites" (Appendix 2-4). The AL including both *Europe/[European] Union* and *Africa* indicate a fixation on the impact of migration on Europe. For example, AL *Africa, Europe, migrant → African* occurs in an article about "Africanization of Europe": "Let us remind ourselves now how Europe looks half African now already, and let us pray that it would not be completely Africanized in a couple of years" (Appendix 2-5).

Images of migrants as different and incompatible with Europe are distinctly present in texts that contain ALs with these KWs. For example, an article containing AL *migration, migrant, union* ➞ *Africa* represents migrants as a natural disaster ("flooding") and enemies ("ruin[ing] the fortress of Europe") and as "a black continent becoming a demographic atomic bomb" (Appendix 2-6).

**Czech government's collusion with the EU**
ANTS articles blame the leading Czech politicians for being subservient to the EU, to the detriment of Czechs. These views are circulated in texts with ALs that contain KWs connected with the redistribution of migrants, the "politics" of migration, and countries that support or oppose such policies.

An article with the top AL *Dublin, migration, migrant* ➞ *IV* accuses the current Czech government as lying to its public: "[Prime minister] Babiš's government in Luxembourg again did not oppose Dublin IV at the EU Council of Ministers, […] ANO 2011 [the leading political party] says something else at home, but it keeps close pace with Berlin in voting in the EU! […] Something stinks!" (Appendix 2-7). Similarly, the AL *migration, migrant, politics* ➞ *migration (adj)* occurs in the following article to suggest that the government is lying: "The Czech government and the prime minister, wherever he goes, he claims that the Czech migration politics is identical to the Hungarian approach in this area. Why, then, is he taking his time and not taking action?" (Appendix 2-8)

The Czech government is contrasted with other governments that support or oppose acceptance of migrants. Hungary is a KW that is often used to create such a contrast, in the following text containing the AL *Hungary, migrant* ➞ *Hungarian*: "Migration from Africa to the EU is to continue in full swing: What didn't go through the door, crawls in through the window. We will go ahead and invite them all here! We are fed lies that the invasion is easing. Only Hungary didn't sign, and we [say] YES [simultaneously the name of the leading party ANO]). How to cheat the public quietly?" (Appendix 2-9). The text suggests migrants are sneaky and invasive ("crawls in through the window," "invasion"), the EU is irresponsible ("We will go ahead and invite them all here!"), and the Czech government is cheating the public. In comparison to Hungary, Germany that plays leading roles in the EU, is presented as being internally in disarray, such as with the AL *migrant, Germany, German* ➞ *Merkel,* which is found in a text about the German coalition's breaking up: "In Germany, the governing coalition of the CDU and CSU is breaking up over Dublin IV, with Horst Seehofer wanting to return migrants from the border from July 1, regardless of whether the EU Council votes to reform Dublin in late June." (Appendix 2-10).

**EU as an authoritarian system and a need for Czech EU-exit**
Another discernible narrative concerns a stance in favor of the Czechs exiting the EU, which is especially visible in texts with ANTS-unique KWs referring to the EU ([European] *Union*, [European] *commission*, *member* [states]) and the Czech Republic (*CzR*, *minister* of *interior* [responsible for the country's security], and the possessive pronoun *our*). For example, the AL *Africa, migrant, union* ➞ *migration* occurs in a text about the Marrakesh declaration. The author claims that the declaration "confirms that EU membership does not exist without the acceptance of migrants," and laments that only Hungary has anti-migration politicians who "defend the national interests." The text suggests that EU-exit is the only way for member states to protect their national interests (Appendix 2-11). Additionally, the EU is said to be directly responsible for the migration crisis in a text containing the AL *CzR, migrant, country* ➞ *member*: "What is the main reason for the influx of immigrants to Europe? It is the EU itself!" (Appendix 2-12). ANTS, moreover, creates an image of the EU as an undemocratic organization. For example, the AL *commission, migrant, country* ➞ *member (adj)* occurs in a text stating that the European commission "announced acceptance of migrants on its own behalf of all the EU member states" although "both Hungary and Poland objected to the announcement" (Appendix 2-13).

The argument for Czech EU-exit is further expressed in a text containing the AL *migrant, our, country ➝ member*. According to the author of the text, "defending the borders from migrants without leaving the EU at the same time" as "erroneous", because the latter "is organizing the move." By presenting the acceptance of migrants and "becoming part of the European Islamic Caliphate" in a cause-effect relationship, the author suggests Czech EU-exit, since EU membership would lead to the destruction of the nation (Appendix 2-14). Another text with the AL *migrant, minister ➝ interior* claims the contradictory actions by the pro-EU Czech government ("the new interior minister is disseminating manuals to municipalities to allow foreigners to be included on the electoral lists and candidates for the fall elections" while the media reports "that the government does not want migration") and argues: "Foreigners will flow into the Czech Republic regardless of what is said and promised to voters in the media. This can be stopped only by exiting the EU and NATO." (Appendix 2-15).

**The mainstream media hiding the truth (in contrast to ANTS)**
Texts where *migrant* is keyed simultaneously serve to discredit the Czech mainstream media. The KWs connected with official mechanisms of migration and declarations (*applicant, asylum, document, legal, declaration*) help create an impression that ANTS scrutinizes migration much more rigorously than CR. For example, a text containing AL *migrant, applicant, country ➝ asylum* states that the Czech media is silent when it should signal a "high alert" (like ANTS is) when the EU parliament started discussing "unilateral" redistribution of migrants (Appendix 2-16).

The AL *migrant, refugee, country ➝ asylum* is in a text that claims to bring a thorough scrutiny of the Marrakesh Declaration, which was signed to defend the rights of religious minorities in predominantly Muslim countries. The outlet claims to present the whole truth to the audience about migrants and refugees in contrast to the other media outlets that are hiding some devastating information by bringing "as the apparently first server in the Czech Republic the full version and the Czech translation of the 'scandalous and explosive Marrakesh Declaration'" (Appendix 2-17). A text containing the ALs *document, migration, migrant ➝ legal* and *declaration, migrant ➝ legal* attacks the mainstream media, which previously criticized Aeronet (one of the ANTS outlets) of spreading disinformation about the document of the Marrakesh Declaration by belittling those who believe the Czech mainstream TV: "Who is illiterate must rely on the 'truth' from Kavčí hory [the Czech TV headquarters]" (Appendix 2-18).

**Global conspiracy by transnational organizations**
ALs with KWs that refer to global organizations such as the AL *migration, migrant, UN ➝ legal* are found in an article hinting at no future for the EU because it continues to organize and finance the UN plan: "The European Union cannot be reformed. The EU's bureaucratic apparatus cannot be replaced. The EU is still on its migration plans and the UN's migration plans. Step by step, the EU is implementing, organizing and generously supporting these plans." (Appendix 2-19) Moreover, an example of the AL *global, migrant, UN ➝ migration* indicates that ANTS connects the migrant crisis to a global conspiracy: "a group of pro-UN/EU oligarchs is using the EU tax revenues to promote its own goals up to the point of complete dissolution of the national welfare states in the EU in the migration chaos" (Appendix 2-20). Soros, a billionaire investor-philanthropist, is said to be one of the oligarchs: The [UNHCR] provision is said to "give the green light to thus far hidden sponsors of migration and migration NGOs – to oligarchs of the Soros type!" (Appendix 2-21).

A more full-blown narrative about such a global conspiracy is found in texts that contain *NGO*, such as in the AL *EU, migrant, NGO ➝ migration* (Appendix 2-22): "[…] Arabic billboards started appearing in Prague! This is a blatant provocation of the nonprofit organization People in Need! The American neocons and their NATO are behind the migration to the EU!" The text advocates not only Czech EU-exit, but also NATO-exit, since the massive migration was said to have started and is being directed by Americans from 2011 with the help of NATO. According to this narrative, "Islamic migration" is "a massive plan by American neocons to break up the EU and prevent a globalist link between Berlin and Moscow" by driving the Central European countries to exit the EU because of migration and to "fall under the influence of NATO as a striking fist against Russia."

### 4.2.3. Summary

This section compared lift outliers in CR and ANTS with special attention to those *migrant*-related ALs which contained KWs that were unique to each media class. The ALs and the texts containing ALs indicate differences in the narratives in which the KW *migrant* is embedded. The CR outliers point to texts with a compact narrative focusing on descriptions of the difficult EU debates on migrants.

The ANTS outliers, in contrast, point to several narrative lines: the destructive nature of immigration organized by the pro-EU Czech government, the EU and globalists and the mainstream media as misleading the public. These stories are used to support a Czech exit from the EU and NATO, possibly to avoid a confrontation with Russia suggested in some of the texts. The outlier ALs in ANTS clearly demonstrate that the media class, by virtue of its anti-migration stance, severely denies the human aspects of migration (e.g., tragedies, sufferings, and integration of migrants). Although this is a pilot study based on one seed KW *migrant*, this section serves as an illustration of how much insight is obtainable from ALs involving one KW from MBA about the differences between the two media classes, especially by identifying the predominant narratives within them.

## 5. Conclusions

Traditionally, corpus-based discourse analysis entails KW extraction followed by various methods (collocation analysis, close reading of concordance lines, etc.), which facilitate the interpretation of KWs by "re-contextualizing" them. These follow-up methods allow researchers to use keyword analysis as a way to focus on a manageable number of the potentially most relevant discourse items. However, as the target of analysis grows increasingly large, it becomes important to find a systematic method not only to downsize the number of KWs but also to identify the most relevant texts. Moreover, as argued above, the existing post-KW-extraction methods do not have the capacity to address all three major features (frequency, dispersion and context) needed to interpret KWs. Among these three, information on context is the area where MBA could contribute most as a post-KW-extraction method.

MBA, which has built-in evaluative measurements (namely support and lift), is a different type of post-KW-extraction method. It enables the identification of the KWs which systematically co-occur in the texts of a corpus, thereby informing us of not only the wider context in any one text, but also of the entire set of texts in which a KW appears in relation to other KWs. By taking into account all occurrences of a KW in all texts in the corpus, MBA informs about the types of prominent concepts that tend to regularly co-occur in the discourse. MBA, in combination with examination of texts where ALs occur, can generate cues to what dominant narrative lines are unique to a media class/corpus, and as the pilot study on anti-system media showed, how these narrative lines lead to an overarching and persistent story of the media class, revealing some systematic features of what is called the "flooding" strategy of random concepts.

# Acknowledgments

The authors would like to thank the three anonymous referees for their productive comments. We would also like to thank Prof. Laura Janda for encouraging us to present the idea, and Andrew Malcovsky for his careful copyediting. Any errors and inconsistencies that remain are of course the authors' responsibility.

This work was supported by the European Regional Development Fund project "Creativity and Adaptability as Conditions of the Success of Europe in an Interrelated World" (reg. no.: CZ.02.1.01/0.0/0.0/16_019/0000734). The research for this paper was also partially funded by the Norweigian Research Council no. 300002.

# References

Agrawal, R. and R. Srikant. 1994. 'Fast Algorithms for Mining Association Rules', *Proceedings of the 20th VLDB Conference*, pp. 487–499.

Baker, P. 2005. *Public discourses of gay men*. London: Routledge.

Baker, P. 2006. *Using Corpora in Discourse Analysis*. London/New York: Continuum.

Baker, P. and T. McEnery. 2005. 'A corpus-based approach to discourses of refugees and asylum seekers in UN and newspaper texts', *Journal of Language and Politics 4*(2), pp. 197–226.

Baker, P. and T. McEnery (eds.). 2015. *Corpora and Discourse Studies*. London: Palgrave Macmillan.

Baker, P., C. Gabrielatos, M. KhosraviNik, M. Krzyzanowski, T. McEnery, and R. Wodak. 2008. 'A useful methodological synergy? Combining critical discourse analysis and corpus linguistics to examine discourses of refugees and asylum seekers in the UK press', *Discourse & Society* 19(3), pp. 273–306.

Bertels, A. and D. Speelman. 2013. '"Keywords Method" versus "Calcul des Spécificités": A comparison of tools and methods', *International Journal of Corpus Linguistics*, *18*(4), pp. 536–560.

Church, K. W. and P. Hanks. 1990. 'Word association norms, mutual information, and lexicography', *Computational Linguistics 16*(1), pp. 22–29.

Cruse, D. A. 1986. *Lexical semantics*. Cambridge: Cambridge University Press.

Culpeper, J. 2002. 'Computers, language and characterisation: An analysis of six characters in Romeo and Juliet' in U. Melander-Marttala, C. Ostman, and M. Kyto (eds.) *Conversation in Life and in Literature: Papers from the ASLA Symposium* (Vol. 15), pp. 11–30. Uppsala: Association Suédoise de Linguistique Appliquée.

Culpeper, J. and J. Demmen. 2015. 'Keywords' in D. Biber and R. Reppen (eds.) *The Cambridge Handbook of English Corpus Linguistics*, pp. 90–105. Cambridge: Cambridge University Press.

Cvrček, V. and M. Fidler. 2019. 'More than keywords: Discourse prominence analysis of the Russian Web portal Sputnik Czech Republic.' in M. Berrocal and A. Salamurović (eds.) *Political Discourse in Central, Eastern and Balkan Europe*, pp. 93–117. Amsterdam/Philadelphia: John Benjamins.

Dunning, T. 1993. 'Accurate Methods for the Statistics of Surprise and Coincidence', *Computational Linguistics 19*(1), pp. 61–74.

Egbert, J. and P. Baker. 2019a. *Using corpus methods to triangulate linguistic analysis*. New York / London: Routledge.

Egbert, J. and D. Biber. 2019b. 'Incorporating text dispersion into keyword analyses', *Corpora 14*(1), pp. 77–104.

Fidler, M. and V. Cvrček. 2015. 'A Data-Driven Analysis of Reader Viewpoints: Reconstructing the Historical Reader Using Keyword Analysis', *Journal of Slavic Linguistics 23*(2), pp. 197–239.

Fidler, M. and V. Cvrček. 2017. 'Keymorph Analysis, or How Morphosyntax Informs Discourse', Corpus Linguistics and Linguistic Theory 15(1), pp. 39-70.

Fidler, M. and V. Cvrček. 2018. 'Going Beyond "Aboutness": A Quantitative Analysis of Sputnik Czech Republic' in M. Fidler and V. Cvrček (eds.) *Taming the corpus: From inflection and lexis to interpretation,* pp. 195–225. Cham: Springer.

Fidler, M. and V. Cvrček. 2019. 'Keymorph analysis, or how morphosyntax informs discourse', *Corpus Linguistics and Linguistic Theory*, *15*(1), pp. 39–70.

Fidler, M. and V. Cvrček. 2020. 'Anti-system Web Portals and their Network of Meaning: A Corpus-based Approach in Czech'. Paper presented at the annual national conference of the American Association of Slavic and East European Languages (available at https://www.brown.edu/research/projects/needle-in-haystack/sites/brown.edu.research.projects.needle-in-haystack/files/uploads/AATSEEL2020-v5-muf.pdf)

Fischer-Starcke, B. 2009. 'Keywords and frequent phrases of Jane Austen's Pride and Prejudice', *International Journal of Corpus Linguistics*, *14*(4), pp. 492–523.

Gabrielatos, C. and P. Baker. 2008. 'Fleeing, Sneaking, Flooding: A Corpus Analysis of Discursive Constructions of Refugees and Asylum Seekers in the UK Press, 1996-2005', *Journal of English Linguistics* 36(1), pp. 5-38.

Gabrielatos, C., & Marchi, A. (2012). '*Keyness: Appropriate metrics and practical issues*'. CADS International Conference 2012, University of Bologna, Italy.

Gries, S. T. (2008): Dispersion and adjusted frequencies in corpora. *International Journal of Corpus Linguistics*. 13(4), 403–437.

Hahsler, M., C. Buchta, D. Gruen, and K. Hornik. 2019. *arules: Mining association rules and frequent itemsets*. Available at: https://CRAN.R-project.org/package=arules

Hahsler, M., S. Chelluboina, K. Hornik, and C. Buchta. 2011. 'The arules r-package ecosystem: Analyzing interesting patterns from large transaction datasets', *Journal of Machine Learning Research 12*, pp. 1977–1981.

Han, J., M. Kamber, and J. Pei. 2011. *Data Mining: Concepts and Techniques* (3rd edition). Haryana/Burlington: Morgan Kaufmann.

Hofland, K. and S. Johansson. 1982. *Word frequencies in British and American English*. Bergen: Norwegian computing centre for the Humanities.

Information Resources Management Association. 2014. *Computational Linguistics: Concepts, Methodologies, Tools, and Applications* (1 edition). Hershey: IGI Global.

Jelínek, T. 2008. 'Nové značkování v Českém národním korpusu', *Naše řeč*, *91*(1), pp. 13–20.

Křen, M., V. Cvrček, T. Čapka, A. Čermáková, M. Hnátková, T. Jelínek, D. Kováříková, V. Petkevič, P. Procházka, M. Škrabal, P. Truneček, P. Vondřička, and A. J. Zasina. 2015. *SYN2015: Reprezentativní korpus psané češtiny*. Praha: Ústav Českého národního korpusu FF UK. (Available at: www.korpus.cz)

Křen, M., V. Cvrček, T. Čapka, A. Čermáková, M. Hnátková, T. Jelínek, D. Kováříková, V. Petkevič, P. Procházka, M. Škrabal, P. Truneček, P. Vondřička, and A. J. Zasina. 2016. 'SYN2015: Representative Corpus of Contemporary Written Czech', *Proceedings of the Tenth International Conference on Language Resources and Evaluation*, pp. 2522–2528. Portorož: ELRA

Mahlberg, M. 2007. 'Clusters, key clusters and local textual functions in Dickens', *Corpora 2*(1), pp. 1–31.

Miner, G. 2012. *Practical text mining and statistical analysis for non-structured text data applications.* Waltham: Academic Press.

Partington, A. and J. Morley. 2004. 'From frequency to ideology: investigating word and cluster/bundle frequency in political debate', *Practical applications in language and computers*, pp. 179–192.

Petkevič, V. 2014. 'Problémy automatické morfologické disambiguace češtiny', *Naše řeč 4–5*, pp. 194–207.

Petráčková, V., J. Kraus J. et al. 1998. *Akademický slovník cizích slov*. Prague: Academia.

Popescu, I.-I., K.-H. Best, and G. Altmann. 2007: 'On the dynamics of word classes in texts', *Glottometrics* 14, pp. 58–71.

R Core Team. 2018. *R: A Language and Environment for Statistical Computing*. R Foundation for Statistical Computing. (Available at: https://www.R-project.org/)

Reisigl, M. and R. Wodak. 2016. 'The discourse-historical approach (DHA)' in R. Wodak and M. Meyer (eds.) *Methods of Critical Discourse Studies*, pp. 23–61. Los Angeles, London, Ne Delhi, Singapore, Washington DC: Sage.

Scott, M. 2010. 'Problems in investigating keyness, or clearing the undergrowth and marking out trails…' in M. Bondi and M. Scott (eds.) *Keyness in Texts*, pp. 43–58. Amsterdam/Philadelphia: John Benjamins.

Scott, M. and C. Tribble. 2006. *Textual Patterns: Key words and corpus analysis in language education*. Philadelphia: John Benjamins.

Stubbs, M. 1996. Text and Corpus Analysis: Computer Assisted Studies of Language and Culture. Oxford/Cambridge, Mass: Blackwell.

Stubbs, M. 2005. 'Conrad in the computer: Examples of quantitative stylistic methods', *Language and Literature 14*(1), pp. 5–24.

Tabbert, U. 2015. *Crime and Corpus.* Amsterdam/Philadelphia: John Benjamins.

Walker, B. 2010. 'Wmatrix, key-concepts and the narrators in Julian Barnes' Talking It Over' in D. McIntyre and B. Busse (eds.) *Language and Style*, pp. 364–387. Basingstoke: Palgrave MacMillan.

# Appendix 1: Lists of original Czech KWs

| ranking | lhs | rhs | lift | count |
| --- | --- | --- | --- | --- |
| 1 | evropský, migrant, **summit** | migrační | 81.5 | 38 |
| 2 | migrant, **summit** | migrační | 76.2 | 38 |
| 3 | **italský**, migrant | **loď** | 64.6 | 39 |
| 4 | EU, evropský, migrant | migrační | 60.4 | 42 |
| 5 | evropský, migrant, země | migrační | 58.0 | 46 |
| 6 | **italský**, migrant | **Itálie** | 58.0 | 54 |
| 7 | **loď**, migrant | **Itálie** | 57.0 | 42 |
| 8 | evropský, migrant | migrační | 54.8 | 56 |

**Table 3.** Outlier ALs that include *migrant* in the center-right web media, ranked by lift. The "count" column shows the number of *texts* in which the AL is attested.

| ranking | lhs | rhs | lift | count |
| --- | --- | --- | --- | --- |
| 1 | **Dublin**, **migrace**, migrant | IV | 137.1 | 28 |
| 2 | **IV**, migrant | **Dublin** | 127.6 | 34 |
| 3 | EU, **IV**, migrant | **Dublin** | 127.6 | 29 |
| 4 | **IV**, **migrace**, migrant | **Dublin** | 127.6 | 28 |
| 5 | evropský, **IV**, migrant | **Dublin** | 127.6 | 27 |
| 6 | **IV**, migrant, země | **Dublin** | 127.6 | 26 |
| 7 | **Dublin**, migrant | **IV** | 111. | 34 |
| 8 | **Dublin**, EU, migrant | **IV** | 107.4 | 29 |
| 9 | **Dublin**, evropský, migrant ' | **IV** | 105.7 | 27 |
| 10 | **Dublin**, migrant, země | **IV** | 104.8 | 26 |
| 11 | migrant, **žadatel**, země | **azyl** | 88.1 | 23 |
| 12 | migrant, **žadatel** | **azyl** | 80.3 | 31 |
| 13 | **azyl**, migrant | **žadatel** | 77.6 | 31 |
| 14 | **azy**l, migrant,země | **žadatel** | 75.9 | 23 |
| 15 | **Afrika**, migrant, země | **africký** | 75.0 | 26 |
| 16 | **Afrika**, EU, migrant | **africký** | 62.2 | 23 |
| 17 | **Maďarsko**, migrant | **maďarský** | 62.0 | 23 |
| 18 | **Afrika**, **Evropa**, migrant ' | **africký** | 60.6 | 28 |
| 19 | **Afrika**, **migrace**, migrant | **africký** | 59.9 | 24 |
| 20 | **maďarský**, migrant | **Maďarsko** | 58.7 | 23 |
| 21 | **Afrika**, migrant | **africký** | 52.3 | 29 |
| 22 | **africký**, **Evropa**, migrant | **Afrika** | 49.6 | 28 |
| 23 | **dokument** | **legální** | 46.6 | 24 |
| 24 | **deklarace**, migrant | **legální** | 46.1 | 23 |
| 25 | **deklarace**, **migrace**, migrant | **legální** | 46.1 | 23 |
| 26 | **africký**, migrant, země | **Afrika** | 46.1 | 26 |
| 27 | **africký**, EU, migrant | **Afrika** | 43.8 | 23 |
| 28 | **africký**, **migrace**, migrant | **Afrika** | 42.5 | 24 |
| 29 | **migrace**, migrant, **OSN** | **legální** | 41.0 | 29 |
| 30 | **africký,** migrant | **Afrika** | 40.3 | 29 |
| 31 | **dokument**, migrant | **legální** | 38.9 | 25 |
| 32 | migrant, **uprchlík** | **azyl** | 36.5 | 24 |
| 33 | migrant, **OSN** | **legální** | 36.1 | 29 |
| 34 | **migrace**, migrant, **politika** | migrační | 35.9 | 24 |
| 35 | **migrace**, migrant, **uprchlík** | **legální** | 33.7 | 26 |
| 36 | **ČR**, migrant, země | **členský** | 33.6 | 28 |
| 37 | **legální**, **migrace**, migrant | migrační | 32.9 | 25 |
| 38 | **komise**, migrant, země | **členský** | 32.2 | 23 |
| 39 | **Afrika**, migrant, země | **legální** | 31.8 | 23 |
| 40 | **Afrika**, **migrace**, migrant | **legální** | 31.1 | 26 |
| 41 | **migrace**, migrační, migrant | **legální** | 30.5 | 25 |
| 42 | migrant, **Německo**, **německý** | **Merkelová** | 29.4 | 30 |
| 43 | EU, **komise**, migrant | **členský** | 28.8 | 24 |

| 44 | migrant, **náš**, země | **členský** | 28.5 | 26 |
| --- | --- | --- | --- | --- |
| 45 | migrant, **ministr** | **vnitro** | 26.2 | 25 |
| 46 | **ČR**, EU, migrant | **členský** | 26.1 | 28 |
| 47 | Afrika, **Evropa**, migrant | **legální** | 25.9 | 25 |
| 48 | **deklarace**, migrant | **migrace** | 25.9 | 31 |
| 49 | evropský, migrant, **OSN** | **migrace** | 25.9 | 31 |
| 50 | **legální**, migrant, **OSN** | **migrace** | 25.9 | 29 |
| 51 | deklarace, **legální**, migrant | **migrace** | 25.9 | 23 |
| 52 | **globální**, migrant, **OSN** | **migrace** | 25.9 | 23 |
| 53 | migrant, **německý** | **Merkelová** | 25.2 | 33 |
| 54 | **Evropa**, **legální**, migrant | **Afrika** | 25.2 | 25 |
| 55 | evropský, **komise** | **členský** | 25.2 | 24 |
| 56 | **Afrika**, **legální**, migrant | **migrace** | 25.0 | 26 |
| 53 | migrant, **německý** | **Merkelová** | 25.2 | 33 |
| 54 | **Evropa**, **legální**, migrant | **Afrika** | 25.2 | 25 |
| 55 | evropský, **komise** | **členský** | 25.2 | 24 |
| 56 | **Afrika**, **legální**, migrant | **migrace** | 25.0 | 26 |
| 57 | **dokument, legální**, migrant | **migrace** | 24.8 | 24 |
| 58 | **Afrika**, migrant, **unie** | **migrace** | 24.8 | 24 |
| 59 | EU, migrant, **neziskovka** | **migrace** | 24.8 | 24 |
| 60 | migrant, **neziskovka**, země | **migrace** | 24.8 | 23 |
| 57 | **dokument, legální**, migrant | **migrace** | 24.8 | 24 |
| 58 | **Afrika**, migrant, **unie** | **migrace** | 24.8 | 24 |
| 59 | EU, migrant, **neziskovka** | **migrace** | 24.8 | 24 |
| 60 | migrant, **neziskovka**, země | **migrace** | 24.8 | 23 |
| 61 | migrant, **OSN, uprchlík** | **migrace** | 24.8 | 23 |
| 62 | migrace, migrant, **unie** | **Afrika** | 24.7 | 24 |

**Table 4.** Outliers in anti-system web media that include *migrant*, ranked by lift. The "count" column shows the number of *texts* in which the AL is present.

# Appendix 2: List of text segments quoted in the main text in the original

1. Loď s migranty nikdo nechce, Merkelová rebely summitem neuchlácholila (https://www.idnes.cz/zpravy/zahranicni/merkelova-migrace-soder-nemecko-salvini-italie-libye.A180625_144148_zahranicni_aha)

2. Vyplula další loď s migranty. Itálie ji odmítla, nakonec zakotví ve Španělsku (https://www.idnes.cz/zpravy/zahranicni/lod-migranti-italie-proactiva-open-arms-matteo-salvini.A180630_164311_zahranicni_amu)

3. Italské volby ve znamení migrace. V průzkumech vedou populisté, slibující její zastavení, EU ani eurozónu ale nezpochybňují (https://archiv.ihned.cz/c1-66064120-italske-volby-ve-znameni-migrace-v-pruzkumech-vedou-populiste-slibujici-jeji-zastaveni-eu-ani-eurozonu-ale-nezpochybnuji)

4. Naprostá většina našich občanů tady afroislámské násilnické migranty a ekonomické parazity nečeká, nechce a nepotřebuje, s výjimkou skupiny politiků přisluhujících Bruselu a Berlínu a početných neziskovkářů (http://www.rukojmi.cz/clanky/zahranicni-politika/5168-sex-s-imigrantem-voni-po-jasminu-tvrdi-zapadoevropske-feministky-reakce-martina-kollera-je-skutecne-brutalni)

5. Připomeňme si nyní, jak již vypadá již napůl africká Evropa dnes, a modleme se, aby za pár let nebyla afrikanizovaná zcela (http://www.rukojmi.cz/clanky/zahranicni-politika/5051-diktat-volneho-trhu-je-v-africe-nutnou-podminkou-pro-zavedeni-zapadni-demokraticke-diktatury-v-duchu-hesla-valka-je-dobry-byznys-nutne-je-vse-zlikvidovane-znovu-vybudovat-za-pomoci-darcu-a-veritelu-pod-zastitou-mmf-a-svetove-banky)

6. Pobořená pevnost Evropa: Milióny se šikují před branami. Zaplaví Afričané náš světadíl? Neschopnost politiků volá do Nebe. Černý kontinent demografickou atomovkou? Pořádné drama teprve začne https://www.protiproud.cz/politika/3528-poborena-pevnost-evropa-miliony-se-sikuji-pred-branami-zaplavi-africane-nas-svetadil-neschopnost-politiku-vola-do-nebe-cerny-kontinent-demografickou-atomovkou-poradne-drama-teprve-zacne.htm

7. Babišova vláda v Lucemburku se opět na Radě ministrů EU nepostavila proti Dublinu IV, proti bylo jenom 7 zemí spolu s Itálií a Slovenskem, Česko mezi nimi nebylo. Hnutí ANO něco jiného říká doma, ale při hlasováních v EU drží těsný krok s Berlínem! […] Něco tady smrdí! (https://aeronet.cz/news/babisova-vlada-v-lucemburku-se-opet-na-rade-ministru-eu-nepostavila-proti-dublinu-iv-proti-bylo-jenom-7-zemi-spolu-s-italii-a-slovenskem-cesko-mezi-nimi-nebylo-hnuti-ano-neco-jineho-rika-doma/)

8. "Česká vláda a premiér kudy chodí, tudy tvrdí, že česká migrační politika je identická s maďarským přístupem v této oblasti. Proč tedy vyčkává a nekoná?" (https://ceskoaktualne.cz/2018/09/zpravy-k-zamysleni/babisova-vlada-chranit-lidska-prava-vsech-uprchliku-migrantu-bez-ohledu-status-najit-novy-domov-vsechny-uprchliky-potrebne-presidleni)

9. Migrace z Afriky do EU má naplno pokračovat: Co nešlo dveřmi, vleze oknem. Pozveme si je sem všechny! Lžou nám, že se invaze zmírňuje. Pouze Maďarsko nepodepsalo, my ANO. Jak potichu oklamat veřejnost? (https://ceskoaktualne.cz/2018/06/uprchlici-zpravy/migrace-afriky-eu-ma-naplno-pokracovat-neslo-dvermi-vleze-oknem-pozveme-si-sem-vsechny-lzou-nam-ze-se-invaze-zmirnuje-madarsko-nepodepsalo-ano-potichu-oklamat-verejnost/)

10. V Německu se kvůli Dublinu IV rozpadá vládní koalice CDU a CSU, Horst Seehofer chce od 1. července vracet migranty z hranic bez ohledu na to, jestli Rada EU koncem června odhlasuje reformu Dublinu. Vadí mu kamiony, které vozí do Německa migranty z tranzitních zemí, mezi nimi je i Česko! […] (https://aeronet.cz/news/v-nemecku-se-kvuli-dublinu-iv-rozpada-vladni-koalice-cdu-a-csu-horst-seehofer-chce-od-1-cervence-vracet-migranty-z-hranic-bez-ohledu-na-to-jestli-rada-eu-koncem-cervna-odhlasuje-reformu-dublinu/)

11. Neexistuje členství v EU bez přijímání migrantů, potvrzuje Marakešské prohlášení podepsané evropskými ministry v rámci tzv. Rabatského procesu. V prohlášení se mluví o "podpoře rozvojových výhod migrace" a když to viděl maďarský ministr, doslova se zděsil a odmítl dokument podepsat, na rozdíl od českého ministra vnitra.

(https://aeronet.cz/news/neexistuje-clenstvi-v-eu-bez-prijimani-migrantu-potvrzuje-marakesske-prohlaseni-podepsane-evropskymi-ministry-v-ramci-tzv-rabatskeho-procesu-v-prohlaseni-se-mluvi-o-podpore-migrace/)

12. Co je hlavním důvodem přílivu imigrantů do Evropy? Je to EU samotná! http://czechfreepress.cz/tema-tydne/co-je-hlavnim-duvodem-prilivu-imigrantu-do-evropy-je-to-eu-samotna.html

13. Evropská komise si zřejmě není příliš jistá ochotou všech členských států přihlásit se k dodržování Globálního zákona pro migraci. Proto prohlásila, že ho přijme sama za všechny členské země EU. Maďarsko i Polsko se proti tomuto prohlášení ohradily. Česká republika mlčela. (http://www.zvedavec.org/komentare/2018/02/7495-.htm)

14. […] je mylná představa, jak budeme moci bránit naše hranice před migranty, aniž bychom přitom opustili EU, která jejich přísun do Evropy organizuje. […] - zásadním momentem při tom je to, zda: a/chceme: uchovat náš stát, národ, jazyk, způsob života či nikoliv; stát se součástí evropského islámského chálifátu či nikoliv […] " (http://www.novarepublika.cz/2018/02/jak-kam-dale-z-dejinne-krizovatky.html)

15. Cizinci budou do ČR proudit bez ohledu na to, co se říká a slibuje voličům v médiích. Zastavit to jde pouze vystoupením z EU a z NATO, jenže proti obojímu je odpor v české společnosti (https://aeronet.cz/news/cizinci-z-eu-bez-ceskeho-obcanstvi-budou-moci-na-podzim-nejen-volit-ale-dokonce-i-kandidovat-hamackovo-vnitro-rozeslalo-obecnim-uradum-metodicke-pokyny-ve-kterych-doporucuje-volebnim-komisim/)

16. Nejvyšší poplach! Evropský parlament dostal na stůl reformu Dublinského protokolu a česká média mlčí! (http://zvedavec.org/komentare2018/01/7470-nejvyssi-poplach-evropsky-parlament-dostal-na-stul-reformup-dublinskeho-protokolu-a-ceska-media-mlci.htm)

17. Redakce AE News přináší zřejmě jako první server v České republice plné znění a český překlad skandální a výbušné Marakéšské deklarace, kterou počátkem května v Maroku podepsali představitelé evropských a afrických zemí v rámci tzv. Rabatského procesu.[…] Nezapomeňte, že Aeronet je místo, kde už dnes přinášíme zítřek. Bohužel, ten zítřek vypadá jako biblická gehenna. (https://aeronet.cz/news/exkluzivne-cesky-preklad-plneho-zneni-marakesske-deklarace-dokument-ze-ktereho-budete-zvracet-je-to-horsi-nez-jsme-si-mysleli-a-pouze-madarsky-ministr-zahranici-se-pod-tento-skandalni-dokument/)

18. Kdo neumí číst, musí spoléhat na "pravdu" z Kavčích hor. Za všechno prý může Aeronet, od kterého informace přebírají politici! A proč by nemohli? (https://aeronet.cz/news/video-podle-redaktorky-ct-neni-marakesska-deklarace-o-rizeni-migrace-a-do-newsroomu-si-pozve-odborniky-ze-serveru-fiat-alternativy-kteri-to-divakum-ideologicky-spravne-vysvetli-kdo-neumi-cist/)

19. Evropskou unii nelze zreformovat. Byrokratický aparát EU nelze vyměnit. EU jede stále podle migračních plánů svých a podle migračních plánů OSN. EU krok za krokem tyto plány naplňuje, organizuje a velkoryse finančně podporuje. https://www.czechfreepress.cz/evropa/eu-sama-organizuje-migraci-z-afriky-do-evropy-zde-jsou-fakta.html

20. Peníze jejich [členských zemí EU] daňových poplatníků jsou skupinou oligarchů za OSN/EU používány na prosazování cílů této skupiny – a to až do úplného rozpuštění národních sociálních států EU v migračním chaosu. (http://www.zvedavec.org/komentare/2018/02/7495-.htm)

21. Toto ustanovení [UNHCR] […] dává volnou ruku dosud utajovaným sponzorům migrace a migračních neziskovek - oligarchům typu Sorose! (http://www.novarepublika.cz/2017/10/evropska-komise-spousti-pilotni.html)

22. […] v Praze se začaly objevovat arabské billboardy! Je to nehorázná provokace neziskovky Člověk v tísni! Za migrací do EU stojí američtí neoconi a jejich NATO! […] Vystoupení z EU a nevystoupení z NATO je cesta do pekel. […] Migrační vlnu spustili a řidí Američané od roku 2011 za pomoci NATO. […] Migrace a odpor lidí vůči Islámu dožene ve středoevropských zemích voliče k nenávisti vůči EU a k přesunu pod americké teze řízení a moci, pod křídla NATO jako americké úderné pěsti proti Rusku. Islámská migrace je tak mohutným plánem amerických neoconů na rozbití EU a pro zabránění globalistického spojení mezi Berlínem a Moskvou. (https://aeronet.cz/news/foto-skoncily-volby-a-v-praze-se-zacaly-objevovat-arabske-billboardy-je-to-nehorazna-provokace-neziskovky-clovek-v-tisni-za-migraci-do-eu-stoji-americti-neoconi-a-jejich-nato/)